%% file: AutoIAD.tex
\documentclass[letterpaper]{article} 
\usepackage[submission]{aaai2026}  
\usepackage{times}  
\usepackage{helvet}  
\usepackage{courier}  
\usepackage[hyphens]{url}  
\usepackage{graphicx} 
\usepackage{natbib}  
\usepackage{caption} 
\usepackage{algorithm}
\usepackage{amsmath}
\usepackage[noend]{algpseudocode}

\usepackage{amssymb}  
\usepackage{pifont}   

\usepackage{tabularx}
\usepackage{ragged2e} 

\usepackage{newfloat}
\usepackage{listings}
\DeclareCaptionStyle{ruled}{labelfont=normalfont,labelsep=colon,strut=off} 
\floatstyle{ruled}
\newfloat{listing}{tb}{lst}{}
\floatname{listing}{Listing}
\title{AutoIAD: Manager-Driven Multi-Agent Collaboration for Automated \\Industrial Anomaly Detection}
\author {
    Dongwei Ji\textsuperscript{\rm 1,2},
    Bingzhang Hu\textsuperscript{\rm 3},
    Yi Zhou\textsuperscript{\rm 1,2 \thanks{Corresponding author}}
}
\affiliations {
    \textsuperscript{\rm 1}School of Computer Science and Engineering, Southeast University, China\\
    \textsuperscript{\rm 2}Key Laboratory of New Generation Artificial Intelligence Technology and Its Interdisciplinary Applications, Ministry of Education, China\\
    \textsuperscript{\rm 3}Hefei CAS Dihuge Automation Co.,Ltd\\
    jidongwei@seu.edu.cn, yizhou.szcn@gmail.com
}

\usepackage{aaai2026}
\usepackage{times}
\usepackage{helvet}
\usepackage{courier}
\usepackage[hyphens]{url}
\usepackage{graphicx}
\usepackage{amsmath}
\usepackage{amssymb}
\usepackage{caption}
\usepackage{subcaption}
\usepackage{algorithm}
\usepackage{algpseudocode}
\usepackage{booktabs}
\usepackage{multirow}
\usepackage{xcolor}
\usepackage{pifont} 

\begin{document}

\maketitle

\begin{abstract}
Industrial anomaly detection (IAD) is critical for manufacturing quality control, but conventionally requires significant manual effort for various application scenarios. This paper introduces AutoIAD, a multi-agent collaboration framework, specifically designed for end-to-end automated development of industrial visual anomaly detection. AutoIAD leverages a Manager-Driven central agent to orchestrate specialized sub-agents (including Data Preparation, Data Loader, Model Designer, Trainer) and integrates a domain-specific knowledge base, which intelligently handles the entire pipeline using raw industrial image data to develop a trained anomaly detection model. We construct a comprehensive benchmark using MVTec AD datasets to evaluate AutoIAD across various LLM backends. Extensive experiments demonstrate that AutoIAD significantly outperforms existing general-purpose agentic collaboration frameworks and traditional AutoML frameworks in task completion rate and model performance (AUROC), while effectively mitigating issues like hallucination through iterative refinement. Ablation studies further confirm the crucial roles of the Manager central agent and the domain knowledge base module in producing robust and high-quality IAD solutions.
\end{abstract}

\begin{links}
    \link{Code}{https://github.com/ji2814/AutoIAD} 
\end{links}

\section{Introduction}

\textbf{Industrial anomaly detection (IAD)} \cite{iad_survey, mmad} plays a crucial role in ensuring product quality and operational efficiency across manufacturing pipelines, such as semiconductor inspection, metal surface analysis, and textile defect detection. These tasks usually require building models which can detect fine-grained visual defects under highly imbalanced data distribution and limited annotation. During the past decade, a range of learning-based methods have emerged to address these challenges, including image-level classifiers \cite{patchcore} and reconstruction-based models such as autoencoders \cite{fastflow}, which achieve strong performance on standard datasets like MVTec AD \cite{mvtecad, mvtecad2}. Despite these advancements, building effective IAD collaboration frameworks still demands substantial \textbf{manual effort} across the entire pipeline from data preprocessing to model selection and training strategy design.

\begin{figure}[t]
    \centering
    \includegraphics[width=1.0\columnwidth]{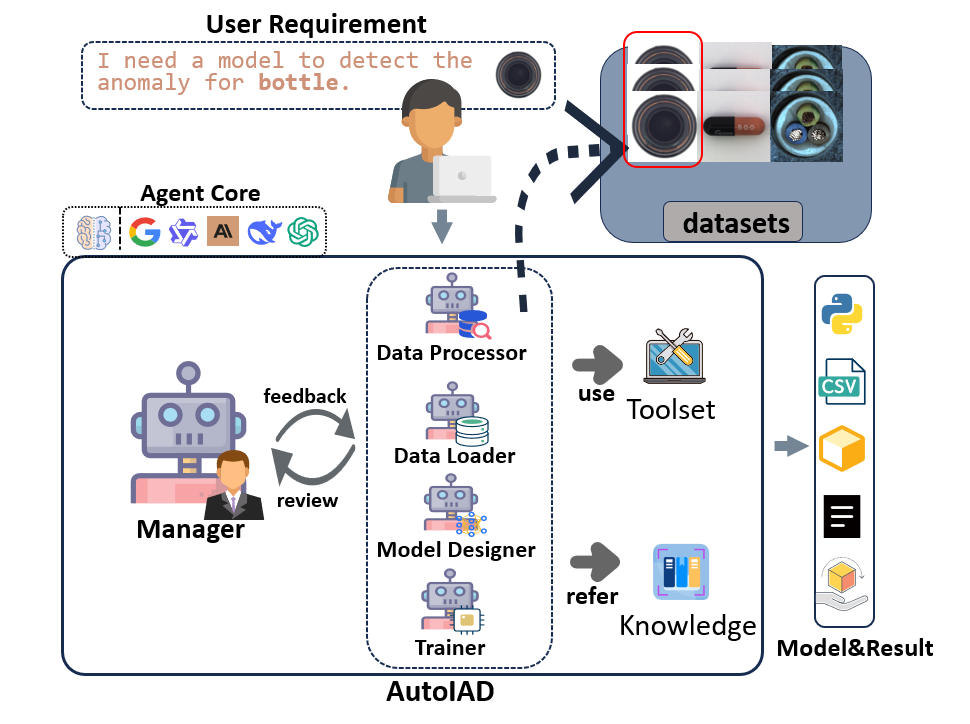}
    \caption{The motivation of the AutoIAD framework for industrial anomaly detection, illustrating the key steps from user requirements to the multi-agents' automated generation of IAD solutions.}
    \label{fig:motivation}
\end{figure}

\begin{table*}[htb]
    \centering
        \begin{tabular}{c|c|c|c|c|c|c}
        \toprule 
        System & Example & Multi-Agent & Manager-driven & Tool Usage & Knowledge Base & Refinement \\
        \midrule 
        ML Agent & ml-agent & \ding{52} & \ding{55} & \ding{52} & \ding{55} & \ding{55} \\ 
        General Agent & openManus & \ding{52} & \ding{55} & \ding{52} & \ding{55} & \ding{55} \\
        AutoML & AutoML-Agent & \ding{52} & \ding{55} & \ding{55} & \ding{55} & \ding{52} \\
        \textbf{AutoIAD} & our work & \ding{52} & \ding{52} & \ding{52} & \ding{52} & \ding{52} \\
        \bottomrule 
        \end{tabular}
    \caption{A comparison of AutoIAD with existing multi-agent frameworks (ML Agent, General Agent, AutoML) regarding multi-agent design, Manager-driven control, tool usage, domain knowledge integration, and optimization capabilities.} 
    \label{tab:comparison}
\end{table*}

More recently, agentic collaboration frameworks powered by large language models (LLMs) \cite{agent_survey} have shown promising performance in automating multi-step workflows for different areas such as autonomous research \cite{research_agent}, software engineering \cite{software_agent}, and financial trading \cite{finance_agent}. These collaboration frameworks coordinate specialized agents under a central planner to decompose and execute complex tasks without relying on human involvement. However, applying this paradigm to industrial vision inspection tasks, particularly IAD, remains underexplored.

Existing LLM-based agents are primarily designed for general-purpose task or code generation, but lack the domain-specific capabilities needed for IAD. They typically do not support automated designing for IAD pipeline. Moreover, these collaboration frameworks often suffer from severe hallucinations \cite{hallucination}, poor inter-agent coordination, and brittle long-horizon execution without human correction.

To bridge this gap, as shown in Figure~\ref{fig:motivation}, we propose \textbf{AutoIAD}, the first multi-agent collaboration framework tailored for industrial anomaly detection. AutoIAD consists of specialized agents, including Data Preparation, Data loader, Model Designer, Trainer, and a central \textbf{Manager} agent that orchestrates the workflow, audits outputs, and supresses hallucination through iterative review. Our framework also integrates a domain-specific knowledge base that encodes augmentation strategies, model heuristics, and tuning practices, where agents can dynamically query to improve decision quality. Unlike generic AutoML and LLM-agent frameworks, AutoIAD enables end-to-end automation from data preparation to finalizing a trained industrial anomaly detection model, which realizes efficient development and iteration of IAD algorithms.

To systematically evaluate AutoIAD, we constructed a comprehensive benchmark designed to assess its performance in automated machine learning (ML) for IAD. This benchmark provides a replicable testbed for future multi-agent collaboration framework for automated IAD research, incorporating evaluation metrics such as task success rate for traditional IAD workflow, latency, token cost, and model performance (e.g., AUROC). Experimentally, we also compare six state-of-the-art (SOTA) LLMs as the agent core for \textbf{AutoIAD}, including claude-3.7-sonnet \cite{claude}, deepseek-v3 \cite{deepseek}, qwen3 \cite{qwen3}.

Our main contributions are highlighted as follows:
\begin{itemize}
\item To our best knowledge, we introduce the first multi-agent collabration framework, \textbf{AutoIAD}, tailored for industrial anomaly detection, featuring a Manager agent that supervises and refines outputs of sub-agents to suppress hallucination and ensure workflow completion rate.
\item We construct a domain-specific benchmark to evaluate LLM agents on industrial anomaly detection tasks, using metrics such as task success rate, execution time, tokens cost and downstream AUROC.
\item Our AutoIAD outperforms existing general-purpose agents and AutoML collaboration frameworks on multiple real-world IAD datasets, setting new standards in automation quality and detection performance.
\end{itemize}

\section{Related Work}

\textbf{Industrial visual anomaly detection} has increasingly adopted unsupervised learning techniques, given the scarcity of labeled defect data and the high diversity of potential anomalies. Classical approaches—such as autoencoders and variational autoencoders—learn to reconstruct normal patterns and identify deviations as anomalies. GAN-based methods like GANomaly \cite{ganomaly} further enhance representation learning through adversarial training. More recently, methods like PaDiM\cite{padim}, PatchCore \cite{patchcore}, and FastFlow \cite{fastflow} utilize pre-trained features and density estimation to achieve state-of-the-art results on benchmarks like MVTec AD \cite{mvtecad, mvtecad2}. However, adapting these methods to new tasks still involves substantial manual effort in data preprocessing, model configuration, and hyperparameter tuning.

\textbf{Large language models (LLMs)} represent a significant leap in AI, pre-trained on vast text data to understand, generate, and process human language \cite{llm, llm-agent}. Their scale enables emergent abilities like in-context learning, adapting to new tasks with minimal examples. Applications are broad: from excelling in natural language processing (NLP) tasks like summarization and question answering, to assisting in code generation and debugging \cite{llm-agent2}. Increasingly, LLMs form the brain of agentic collaboration frameworks, planning and coordinating complex, multi-step workflows by breaking down tasks and interacting with tools.

\textbf{LLM-based agentic collaboration frameworks} have emerged to automate multi-step completing tasks \cite{llm-agentic}. The multi-agent designs simulate collaborative workflows, where specialized AI agents interact to achieve common goals, showing promise in software engineering (e.g., rapid prototyping, code generation). Systems like openManus \cite{openmanus} and openHands \cite{openhands} expand this paradigm to general task planning and completion. Nonetheless, these collaboration frameworks are primarily text-centric or code-centric and are not designed for vision-based ML workflows. They typically lack mechanisms for dynamic model evaluation, iterative refinement, or domain-specific decision making, limiting their applicability in IAD.

\textbf{AutoML frameworks} \cite{autokaggle} aim to reduce manual involvement in model development through pipeline automation. Traditional AutoML frameworks, while effective for general ML tasks, often fall short in IAD due to their lack of support for visual anomaly detection scenarios. For instance, autoMMLab \cite{autommlab} and AutoML-Agent \cite{automl-agent} offer basic capabilities for image classification or object detection, but rely heavily on human intervention and lack dynamic revision or domain-guided augmentation. Moreover, they often do not follow a multi-agent architecture, nor do they feature centralized control for iterative correction or inter-agent feedback.

In contrast, Table~\ref{tab:comparison} compares \textbf{AutoIAD} with representative collaboration frameworks across perspectives such as multi-agent design, manager-driven control, tool usage, domain knowledge integration, and refinement capabilities. AutoIAD, inspired by ml-agent \cite{ml-agent}, is designed specifically for IAD with a dedicated Manager agent to coordinate, validate, and refine the outputs of sub-agents. It incorporates a structured knowledge base that informs domain-specific decisions like augmentation \cite{albumentations}, model selection \cite{anomalib}, and hyperparameter tuning.

\section{Methodology}

\begin{figure}
    \centering
    \includegraphics[width=1.0\columnwidth]{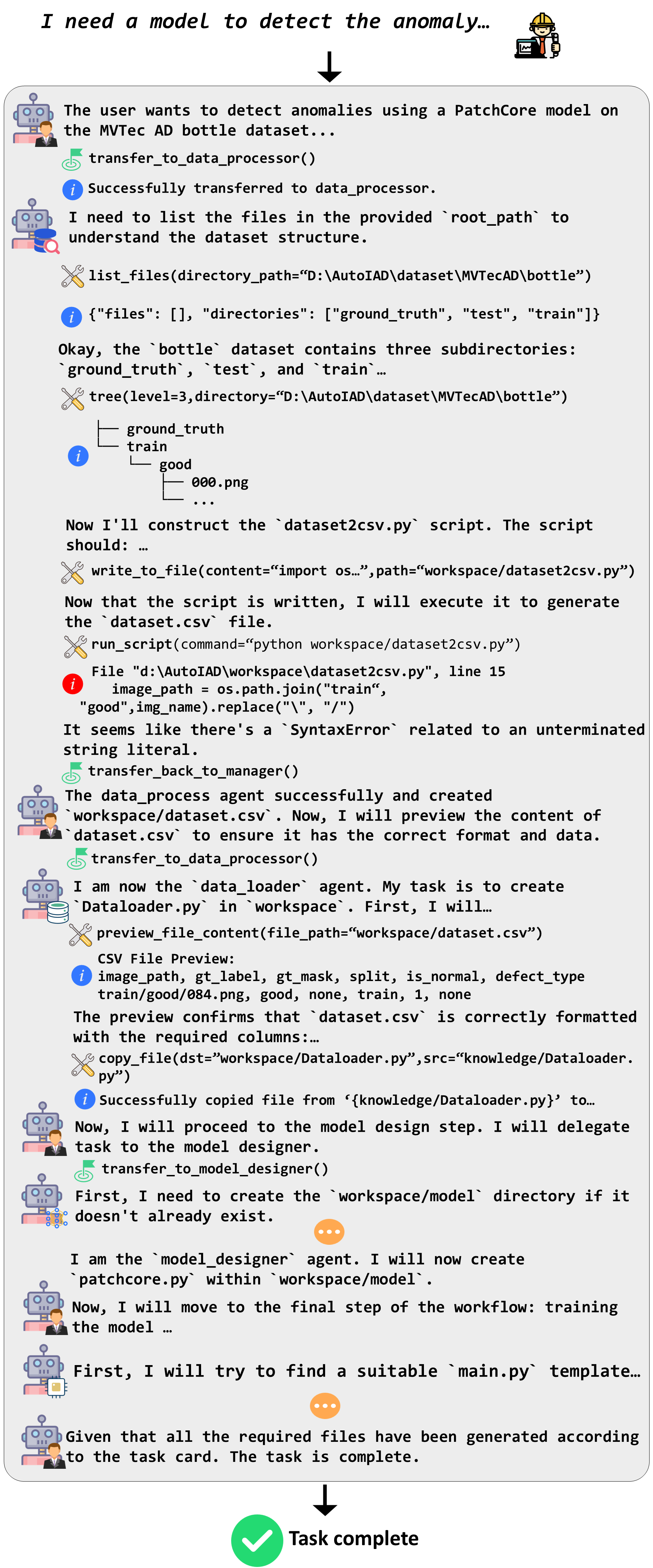}
    \caption{An example of the specific workflow operated by the AutoIAD collaboration framework for IAD tasks, which presents the detailed steps from data file preparation, model design to task completion and the interaction across various agents.}
    \label{fig:workflow}
\end{figure}

In this section, we introduce our multi-agent framework for automated industrial visual anomaly detection. Our framework is designed to autonomously complete the \textbf{end-to-end} pipeline—from dataset preparation, data augmentation to model training and evaluation—through the coordination of specialized agents. Central to our approach is a \textbf{Manager} agent that orchestrates all stages, reviews outputs, and dynamically adjusts the workflow. The collaboration framework also leverages a knowledge module to incorporate industrial expertise, enhancing decision-making throughout the process. An example of the workflow is shown in Figure~\ref{fig:workflow}.

\begin{figure*}[htb]
    \centering
    \includegraphics[width=2.0\columnwidth]{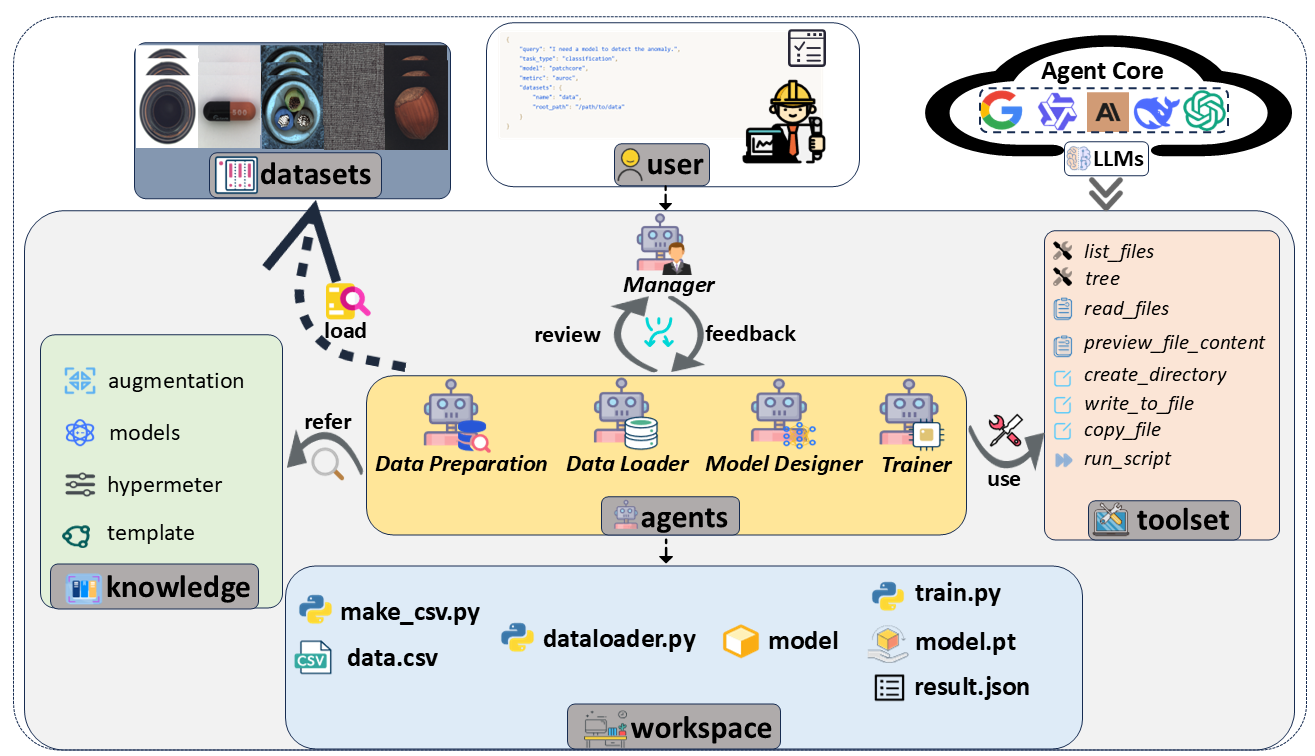} 
    \caption{The overview diagram of our AutoIAD multi-agent framework, showing the collaborative relationship between the Manager Agent and sub-agents including Data Preparation , Data Loader, Model Designer, and Trainer, as well as the interaction with users, datasets, LLMs, toolsets, and knowledge bases.}
    \label{fig:architecture}
\end{figure*}

\subsection{Overview}
We propose \textbf{AutoIAD}, a novel multi-agent framework for the automated development of industrial anomaly detection (IAD) models, depicted in Figure~\ref{fig:architecture}. The collaboration framework supports end-to-end pipeline automation, encompassing data preparation and loading, model design, training, and evaluation. The central innovation lies in the collaborative coordination of specialized autonomous agents, each responsible for a distinct functional stage. This architecture is further empowered by an integrated domain knowledge base and a comprehensive toolset, enabling agents to perceive effectively and execute robust actions in dynamic environments.

\subsection{Multi-Agent Architecture}
The AutoIAD framework is organized around five individual agents, each designed to address a critical task in the IAD pipeline. These agents interact through the central Manager Agent, which orchestrates the workflow, manages contingencies, and ensures the delivery of high-quality anomaly detection models. 

\textbf{Manager Agent.} Within the AutoIAD framework, the Manager Agent serves as the central orchestrator, translating high-level task specifications from the user's requirements into a series of executable subtasks. It then intelligently schedules and delegates these subtasks to specialized worker sub-agents, continuously monitoring their execution progress to ensure a streamlined and efficient workflow.

A crucial function of the Manager Agent is its rigorous quality control. It meticulously validates outputs at each stage, such as the integrity of \texttt{dataset.csv} from the Data Preparation Agent and the functionality of \texttt{Dataloader.py} from the Data Loader Agent. Upon detecting suboptimal performance or errors, the Manager Agent initiates an iterative refinement process, providing targeted feedback to the relevant sub-agents. This sophisticated oversight and iterative correction mechanism is pivotal for mitigating common issues such as hallucination, directly contributing to AutoIAD's demonstrated high task success rates and superior model performance, thereby ensuring a robust and high-quality end-to-end automated solution for industrial anomaly detection.

\textbf{Data Preparation Agent.} The Data Preparation Agent is responsible for transforming raw, industrial image datasets into a standardized, machine-learning-ready format. Its primary objective, orchestrated by the Manager Agent, is to produce a comprehensive \texttt{dataset.csv} file, serving as the canonical input for subsequent data loading and model training phases. The Agent initiates its process by exploring the raw dataset structure, leveraging tools like \texttt{list\_files} and \texttt{tree} to comprehend data organization and select the most suitable data based on specified requirements. Subsequently, it analyzes existing metadata, labels, or other related information to inform its processing logic. This process ensures that the generated \texttt{dataset.csv} provides a consistent, well-structured, and correctly split data foundation, thereby abstracting away the complexities of initial data handling and enabling seamless progression through the automated anomaly detection pipeline.

\textbf{Data Loader Agent.} Within the AutoIAD framework, the Data Loader Agent is tasked with creating a highly efficient and robust \texttt{Dataloader} class, bridging the gap between the standardized \texttt{dataset.csv} generated by the Data Preparation Agent and the machine learning model's training and testing phases. Upon receiving the prepared data, the Data Loader Agent constructs a PyTorch-compatible data loader, encapsulating logic for efficient batch loading across training, validation, and testing stages. It also incorporates mechanisms for randomized splitting and diverse data augmentation strategies to improve the generalization ability of the IAD model. This agent ensures that data feeding mechanisms are reproducible and fully aligned with standard deep learning workflows.

\textbf{Model Designer Agent.} The Model Designer Agent holds the pivotal responsibility of conceiving, selecting or designing, and confirming the suitable core machine learning model for IAD. Drawing from a repository of domain-specific templates, it designs the architecture and validates the configuration for correctness and compatibility. It is also adept at iterative debugging and refinement, guaranteeing the model's robustness and adherence to all specifications. This meticulous process ensures that a well-designed, validated, and task-appropriate model is prepared for subsequent training and evaluation.

\textbf{Trainer Agent.} The Trainer Agent oversees the entire training lifecycle of the model. Its responsibilities include performing hyperparameter tuning, executing training iterations, managing checkpointing, and logging performance metrics. During the post-training stage, this Agent evaluates the model using standard metrics such as AUROC, and communicates the results back to the Manager Agent. In cases of suboptimal performance, the Agent supports retraining with revised configurations, which can ensure optimal model performance.

Collectively, these agents enable a modular, flexible, and fully automated pipeline for IAD model development, substantially reducing the need for manual intervention and domain-specific coding. 

\subsection{Toolset Integration}
To enable autonomous interaction with the operating environment \cite{tools-use}, AutoIAD equips all agents with a curated toolset, which abstracts low-level commands into high-level actions. These tools enable agents to explore over the filesystem, manipulate data, and execute code, thereby bridging the gap between natural language instructions and executable operations.

Key tools integrated in AutoIAD include:
\begin{itemize}
\item \textit{list\_files}, \textit{tree}: Explore and visualize directory structures.
\item \textit{read\_files}, \textit{preview\_file\_content}: Inspect content of datasets and configuration files (e.g., CSV, JSON, TXT).
\item \textit{create\_directory}, \textit{write\_to\_file}, \textit{copy\_file}: Perform essential file system operations.
\item \textit{run\_script}: Execute python or shell scripts for data transformation, model training, or evaluation.
\end{itemize}

These tools form the operational basic units of AutoIAD, empowering agents to dynamically adapt to diverse scenarios and interacting with computer systems. They enable the agents to systematically navigate unknown datasets, intelligently extract critical schema information, and ultimately construct robust and executable pipelines tailored to the specific requirements of industrial anomaly detection tasks.

\subsection{Context-aware Domain Knowledge Reference}
To ensure robustness, efficiency, and task-specific optimization in industrial visual anomaly detection, AutoIAD incorporates a comprehensive \textbf{Domain Knowledge Module}. This module acts as a curated repository of expert-derived priors, heuristics, and templates, empowering agents to make informed, context-aware decisions throughout the entire pipeline. Specifically, it provides:

\begin{itemize}
    \item \textbf{Tailored Data Augmentation Strategies}: This guides the \textit{Data Loader Agent} with techniques like resize, horizontal flip, Gaussian noise, and domain-specific augmentations to enrich datasets and enhance model generalization.
    \item \textbf{Structured Model Definitions and Architectures}: Offering code implementation reference for reconstruction-based, feature embedding-based, and normalized flow models, it also includes guidelines for domain-specific adaptations, enabling the \textit{Model Designer Agent} to intelligently select and configure appropriate models.
    \item \textbf{Standardized Training Scripts and Hyperparameter Optimization Guidance}: Detailing loss functions, evaluation metrics, and offering expert recommendations for critical hyperparameters (e.g., learning rate, coreset sampling ratio, regularization), this streamlines the \textit{Trainer Agent}'s execution of robust training cycles, leading to high-performance anomaly detection models.
\end{itemize}

Agents access this knowledge during critical decision-making, allowing them to make context-aware choices. This integration of domain expertise significantly improves both the pipeline's completion success rate and the performance of the resulting models.

\subsection{Task Specification}

To enable the multi-agent collaboration framework to accurately comprehend and execute IAD tasks, we define the \textbf{TaskCard} as the formalized task description. Each TaskCard precisely delineates critical information pertinent to the designated task, including its type, the intended model name, required evaluation metrics, the dataset's root path, and the specific anomaly category name. A content example of the TaskCard is presented below.
\begin{verbatim}
{
    "query": "I need a model to ...",
    "task_type": "classification",
    "model": "patchcore",
    "metirc": "auroc",
    "datasets": {
        "name": "data",
        "root_path": "path/to/dataset"
    }
}
\end{verbatim}

\begin{algorithm}[htb]
\caption{Overall Procedure of AutoIAD}
\label{alg}
\textbf{Initialization:}
Manager Agent $A_{\text{mgr}}$,
Data Processor Agent $A_{\text{p}}$,
Data Loader Agent $A_{\text{d}}$,
Model Designer Agent $A_{\text{m}}$,
Trainer Agent $A_{\text{t}}$,
system state $S$,
Workspace $W$,
Feedback $F$ and
Output $O$.\\
\textbf{Input:} Task card $T$ \\
\textbf{Output:} Final workspace $W$

\begin{algorithmic}[1] 
    \State \textbf{initialize:}
        $S \gets CONT$,
        $W \gets \emptyset$,
        $A \gets A_{\text{mgr}}$,
        $F \gets \text{None}$,
        $O \gets \text{None}$,
        $Next \gets \text{True}$

    \Function{Call}{agent, W, T, F}
        \State $O \gets \text{agent}(W, T, F)$ \Comment{Agent computation}
        \State $W.\text{agent} \gets O$ \Comment{Write to workspace}
        \State $Next \gets \text{Review}(O)$ \Comment{Self-review}
        \State \Return $Next$
    \EndFunction

    \State $Next \gets \text{True}$
    \While{$S \neq \text{END}$}
        \If{$A = \text{NaN} \lor A = A_{\text{mgr}}$}
            \State $Next \gets \text{True}$
            \State $(A, F, S) \gets A_{\text{mgr}}(W, T)$ \Comment{Schedule agent}
            \If{$S = \text{END}$}
                \State \Return $W$
            \EndIf
        \Else
            \State $agentName \gets \text{Map}(A)$ \Comment{$A_p\rightarrow{}p, A_d\rightarrow{}d, A_m\rightarrow{}m, A_t\rightarrow{}t$}
            \While{$Next$}
                \State $Next \gets \Call{Call}{agentName, W, T, F}$
            \EndWhile
            \State $A \gets A_{\text{mgr}}$
        \EndIf
    \EndWhile
    \State \Return $W$
\end{algorithmic}
\end{algorithm}

\subsection{Procedure of Manager-Driven AutoIAD}
We outline the overall procedure of AutoIAD in Algorithm 1. The process begins with the \textbf{Initialization Stage} (Line 1). Here, all relevant agents are instantiated: Manager Agent ($A_{\text{mgr}}$), Data Preparation Agent ($A_{\text{p}}$), Data Loader Agent ($A_{\text{d}}$), Model Designer Agent ($A_{\text{m}}$), and Trainer Agent ($A_{\text{t}}$). Concurrently, system components are set up: the system state $S$ (which can be \texttt{CONT} for continuation or \texttt{END} for process termination), the Workspace $W$ (serving as a shared repository for agent outputs), the Feedback $F$ (enabling the Manager Agent to refine sub-agent operations), and individual agent outputs $O_i$ (where $i \in \{\text{p, d, m, t}\}$ corresponds to the respective agents). The task card $T$ is provided as the initial input to the system.

In the (2) scheduling and execution stage, the Manager Agent ($A_{\text{mgr}}$) plays a central role. When the condition in Line 9 is met (either the current agent $A$ is not available or is $A_{\text{mgr}}$), $A_{\text{mgr}}$ will schedule agents through its computation in Line 11. The function $\text{CALL}$ (Lines 2 - 6) is used to make an agent perform its computation, write results to the workspace, and conduct self-review to determine the next step. Different agents ($A_{\text{p}}, A_{\text{d}}, A_{\text{m}}, A_{\text{t}}$ etc.) will be called in sequence as needed during the while loop (Lines 8 - 18) to execute tasks related to the input task card $T$, and the workspace $W$ is continuously updated.

Finally, when the system state $S$ reaches $\text{END}$ (Line 12), the final workspace $W$ is returned as the output.

\section{Experiments and Results}

We evaluate \textbf{AutoIAD} on a comprehensive set of Industrial Anomaly Detection (IAD) model development tasks using the widely adopted MVTec AD dataset \cite{mvtecad, mvtecad2}. To rigorously test collaboration framework capabilities, we empower the automation framework with six advanced large language models (LLMs), spanning high-cost and low-cost commercial large language model (LLM) APIs, as well as cutting-edge open-source alternatives. The LLMs include: Claude-3.7-Sonnet \cite{claude}, Qwen-Max \cite{qwen2.5}, Gemini-2.5-Flash-Preview-05-20 \cite{gemini}, GPT-4o-Mini \cite{gpt}, DeepSeek-Chat-v3-0324 \cite{deepseek}, and Qwen3-235B-A22B \cite{qwen3}.

We evaluate the effectiveness of each configuration using multiple metrics: task success rate of the IAD pipeline (retrieve dataset, load data, design model, and train model), execution time, token usage (completion and prompt), and AUROC (Area Under the Receiver Operating Characteristic Curve) for model performance.

\subsection{Experiment Setup}

The evaluation spans 15 carefully defined model-building tasks aligned with MVTec AD's object and texture categories, such as bottle, cable, metal nut, carpet, tile, etc. Each task is governed by a standardized task card, ensuring consistent goals across LLMs.

MVTec AD offers high-resolution images for each category, containing defect-free training data and annotated test sets (with pixel-level anomaly masks). To be consistent with best practices in unsupervised anomaly detection, models are trained exclusively on normal samples and evaluated on both normal and defective test images.

\begin{table*}[th]
    \centering
    \begin{tabular}{c|c|c|c|c|c}
        \toprule 
        Name & Success Rate (\%) & Time (s) & Completion Tokens & Prompt Tokens & AUROC (\%) \\
        \midrule
        \texttt{MLAgent-Bench} & 0 & 123.30 & 28,445 & 4,819 & - \\
        \texttt{AutoML-Agent} & 0 & 318.37 & 77,709 & 17,233 & - \\
        \texttt{openManus} & 50.0 & 103.54 & 141,489 & 6,960 & 48.09 \\
        \texttt{openHands} & 73.3 & 215.2 & 150,009 & 3,559 & 53.88 \\
        \texttt{ours} & \textbf{88.3} & 335.01 & 1,557,258 & 18,797 & \textbf{63.69} \\
        \midrule
        \texttt{w/o manager} & 83.3 & 220.08 & 186,937 & 13,904  & 35.01 \\
        \texttt{w/o knowledge} & 60.0 & 218.21 & 314,342 & 14,277 & 0.0 \\
        \bottomrule 
    \end{tabular}
    \caption{Ablation study and comparison with other agentic collaboration frameworks (using Gemini as core LLM). A dash (-) indicates that the framework failed to complete the full pipeline, thus no AUROC could be computed.} 
    \label{tab:ablation}
\end{table*}

\begin{table*}[th]
    \centering
    \begin{tabular}{c|c|c|c|c|c}
        \toprule 
        LLM & Success Rate (\%) & Time (s) & Completion Tokens & Prompt Tokens & AUROC (\%) \\
        \midrule 
        \texttt{Sonnet3.7} & 63.3 & \textit{300.00} & 284,996 & 10,215 & - \\
        \texttt{Qwen-Max} & 77.8 & 294.55 & 208,119 & 6,412 & 25.71 \\
        \texttt{Gem2.5Flash} & \textbf{88.3} & 335.01 & 1,557,258 & 18,797 & \textbf{63.69} \\ 
        \texttt{GPT-4o-Mini} & 43.3 & 216.98 & 289,907 & 6,462 & 25.00 \\
        \texttt{DeepSeek-v3} & 37.8 & 290.86 & 415,601 & 47,103  & 0.0 \\
        \texttt{Qwen3-235B} & 50.0 & 465.28 & 24,935 & 9,998 & 28.65 \\
        \bottomrule 
    \end{tabular}
    \caption{Performance comparison of different LLM backbones with AutoIAD. In Sonnect3.7, the task was not fully completed, due to exceeding the allocated time limit.} 
    \label{tab:performance}
\end{table*}

\begin{table}[th]
    \centering
    \setlength{\tabcolsep}{1mm}
    \begin{tabular}{c|c|c|c|c|c} 
        \toprule
        & Suc. & Time & Tkns. (I/O) & AUROC (\%) \\ 
        \midrule
        bottle       & 4/4 & 550.2 & 1311445/18574  & 0.0 \\
        cable        & 3/4 & 377.6 & 1728461/25991  & - \\
        capsule      & 3/4 & 153.0 & 368540/14132  & - \\
        carpet       & 4/4 & 495.4 & 855087/17164 & 98.15 \\
        grid         & 3/4 & 158.9 & 1615811/13562 & - \\
        hazelnut     & 4/4 & 170.7 & 312877/16484 & 75.36 \\
        leather      & 3/4 & 127.0 & 2382107/6301 & - \\
        metal\_nut   & 4/4 & 191.9 & 294579/25193 & 85.48 \\
        pill         & 3/4 & 92.9  & 78098/4835  & - \\
        screw        & 4/4 & 313.8 & 246538/11614 & 81.34 \\
        tile         & 4/4 & 577.9 & 5849630/51852 & 89.91 \\
        toothbrush   & 4/4 & 727.4 & 3747066/39741 & 0.0 \\
        transistor   & 4/4 & 261.7 & 2724330/18983  & 79.30 \\
        wood         & 3/4 & 533.2 & 1573120/9357  & - \\
        zipper       & 3/4 & 293.7 & 271186/8174 & - \\
        \bottomrule
    \end{tabular}
    \caption{Experimental Results Summary (Gemini-2.5-Flash-Preview), the metric is success rate (Suc.), times, completion tokens and prompt tokens (Tkns (I/O)), as well as AUROC.}
    \label{tab:results}
\end{table}

\subsection{End-to-End System Comparison}
We further test AutoIAD against other agentic collaboration frameworks: MLAgent-Bench, openManus, openHands and AutoML-Agent, all configured with Gemini as the core.

As shown in Table~\ref{tab:ablation}, AutoIAD significantly outperforms alternatives. MLAgent-Bench fails all tasks due to limited understanding of structured anomaly detection pipelines. openManus achieves a moderate 50.0\% success rate but lacks IAD-specific tuning. Our work, AutoIAD, achieves an 88.3\% success rate and a high AUROC of \textbf{63.69\%}, validating its domain-aware architecture and robust agent collaboration.

\subsection{Task Completion Analysis}
Each LLM is independently run on the 15 tasks, limited to a maximum of 100 recursion steps and a time cap of 300 seconds for high-cost models (claude-3.7-Sonnet and qwen-max), 600 seconds for other alternatives. Task completion is defined as producing a model with acceptable test performance.

As shown in Table~\ref{tab:performance}, performance varies widely across LLMs. Gemini-2.5-Flash-Preview achieves the highest success rate (87.3\%) and the best AUROC (63.69\%), demonstrating its superior code generation and tool orchestration abilities. The detailed experimental results for Gemini-2.5-Flash-Preview are shown in Table~\ref{tab:results}, indicating a high success rate for most MVTec AD categories. While several LLMs (e.g., DeepSeek-v3 and gpt4o-mini) show limited success, suggesting potential gaps in domain-specific action or tool usage capabilities. Claude-3.7-Sonnet achieves a low success rate due to its longer call times.

These results demonstrate that AutoIAD, when powered by well-suited LLMs, can autonomously construct high-performing IAD models across diverse task types.

\subsection{Ablation Study}
To understand the contribution of system components, we conduct ablation experiments. Removing the Manager Agent only slightly reduces the success rate (from 88.3\% to 83.3\%) but drastically degrades performance quality—AUROC drops from 63.69\% to 35.01\%. This underscores the manager’s role in refining decisions and validating outputs. Besides, excluding the IAD knowledge module leads to a significant drop in task success (to 60.0\%) and AUROC (to 0.0\%) which indicates that the pipeline was completed, but the resulting model produced AUROC score that was numerically NaN (Not a Number), confirming that domain knowledge is crucial for correct model setup and training.

\subsection{Resource and Cost Analysis}

\begin{figure}[htb]
    \centering
    \includegraphics[width=1.0\columnwidth]{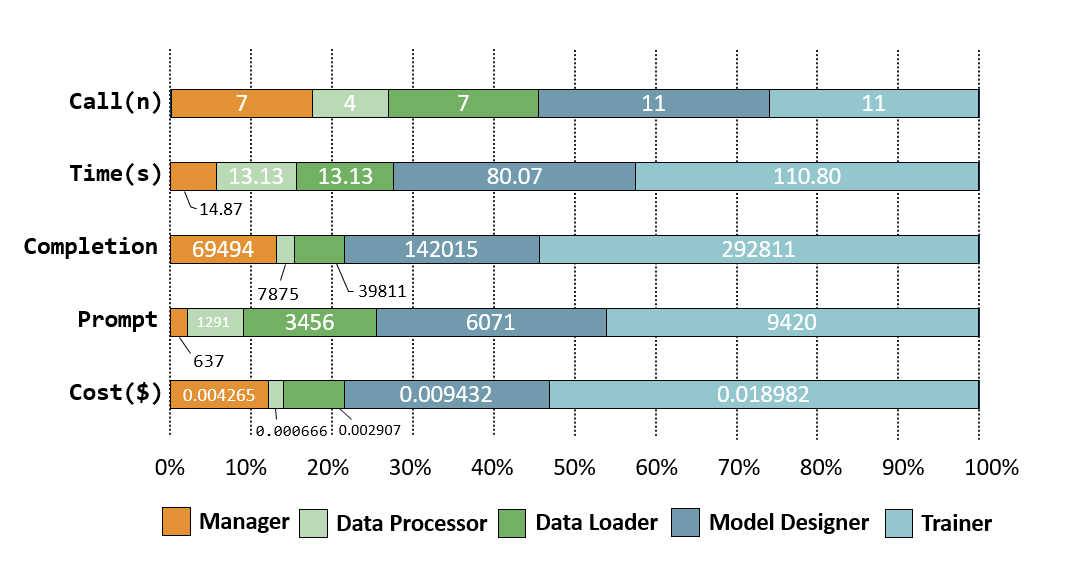}
    \caption{A comparison chart of the number of calls, time consumption, completion tokens, prompt tokens, and cost proportion among agents in the task.}
    \label{fig:case}
\end{figure}

A detailed breakdown of token consumption and execution time by agent roles, as illustrated in Figure~\ref{fig:case}, provides valuable insights into the computational demands of AutoIAD’s pipeline. The Trainer and Model Designer agents exhibit the highest computational cost and time expenditure. This is inherently reflective of their essential and computationally intensive roles in the anomaly detection pipeline, specifically concerning model architecture designing, iterative model training and hyperparameter tuning. However, a noteworthy observation emerges regarding the Manager agent: despite its remarkably low token usage—indicating an efficient operation with minimal conversational overhead—it plays a pivotal role in overall performance curation.

\section{Conclusion}
This paper introduces AutoIAD, a novel multi-agent framework automating industrial anomaly detection (IAD) pipeline development. AutoIAD leverages specialized agents, a central Manager, and a domain-specific knowledge base to reduce manual effort and improve efficiency. Evaluated on MVTec AD, AutoIAD outperforms other collaboration frameworks and sets a new benchmark for automated IAD model development, promising more efficient industrial vision solutions.

\section{Acknowledgements}
This work was supported by the Fundamental Research Funds for the Central Universities (2242025K30024).
\bigskip

\bibliography{aaai2026}

\clearpage
\input{Appendix}
\end{document}

%% file: Appendix.tex
{\centering \bfseries \huge Appendix \par}

\section{Experimentation Details}
This section provides a thorough overview of our experimental methodology, detailing the agent's architectural components, the comparative analysis approach, and the design of the ablation studies.

\subsection{Details of Comparison Experiments}
To rigorously assess the proposed our framework's performance, we conducted comparative experiments against several established baselines. These baselines represent diverse approaches to automated task execution, offering a comprehensive context for evaluating our agent's strengths and limitations.

\subsubsection{Baseline: MLA-Bench}
We benchmark our agent against MLA-Bench, a multi-agent collaboration framework tailored for machine learning tasks. This direct comparison provides insight into our agent's performance relative to a prominent framework in the field. The experimental results are presented in Table~\ref{tab:mla-bench}.

\begin{table*}[h!]
\centering
\caption{Performance metrics for the MLA-Bench baseline. A constant success rate of 0 for all tasks indicates fundamental limitations of this baseline in the evaluated scenarios.}
\label{tab:mla-bench}
\begin{tabular}{lcccc}
\toprule
Task & Success & Time (s) & In/Out Tokens & AUROC \\
\midrule
bottle        & 0/4       & 126.87     & 29270/4959             & NaN   \\
cable         & 0/4       & 105.30     & 24295/4116             & NaN   \\
capsule       & 0/4       & 112.80     & 26024/4409             & NaN   \\
carpet        & 0/4       & 125.69     & 28999/4913             & NaN   \\
grid          & 0/4       & 146.32     & 33758/5719             & NaN   \\
hazelnut      & 0/4       & 122.53     & 28268/4789             & NaN   \\
leather       & 0/4       & 122.97     & 28369/4806             & NaN   \\
metal\_nut    & 0/4       & 129.18     & 29804/5049             & NaN   \\
pill          & 0/4       & 85.67      & 19765/3348             & NaN   \\
screw         & 0/4       & 135.65     & 31296/5302             & NaN   \\
tile          & 0/4       & 160.50     & 37027/6274             & NaN   \\
toothbrush    & 0/4       & 137.19     & 31650/5363             & NaN   \\
transistor    & 0/4       & 116.16     & 26799/4541             & NaN   \\
wood          & 0/4       & 133.24     & 30739/5209             & NaN   \\
zipper        & 0/4       & 89.37      & 20618/3494             & NaN   \\
\bottomrule
\end{tabular}
\end{table*}

\subsubsection{Baseline: openManus}
Our system is evaluated against openManus, an open-source framework designed for general manipulation tasks. This comparison highlights our agent's superior performance in the domain of Intelligent Anomaly Detection (IAD) when compared to frameworks designed for broader task sets. The experimental results are detailed in Table~\ref{tab:openManus}.

\begin{table*}[h!]
\centering
\caption{Performance metrics for the openManus baseline.}
\label{tab:openManus}
\begin{tabular}{lcccc}
\toprule
Task & Success & Time (s) & In/Out Tokens & AUROC \\
\midrule
bottle        & 1/4     & 135.85     & 291232/13462             & NaN   \\
cable         & 2/4     & 101.33     & 127349/6874             & NaN   \\
capsule       & 4/4     & 152.94     & 167206/6383             & 0.4809      \\
carpet        & 1/4     & 70.22      & 70934/1904             & NaN   \\
grid          & 2/4     & 117.02     & 175601/10386             & NaN   \\
hazelnut      & 1/4     & 111.18     & 123098/6893             & NaN   \\
leather       & 0/4     & 99.71      & 135586/5294             & NaN   \\
metal\_nut    & 1/4     & 38.92      & 21034/2800             & NaN   \\
pill          & 3/4     & 129.55     & 99663/6355             & NaN   \\
screw         & 1/4     & 105.73     & 159644/3827             & NaN   \\
tile          & 1/4     & 43.51      & 20527/3372             & NaN   \\
toothbrush    & 2/4     & 108.68     & 161115/9923             & NaN   \\
transistor    & 2/4     & 89.33      & 127084/7681             & NaN   \\
wood          & 3/4     & 137.70     & 269193/12576             & NaN   \\
zipper        & 2/4     & 111.41     & 173067/6674             & NaN   \\
\bottomrule
\end{tabular}
\end{table*}

\subsubsection{Baseline: openHands}
This section presents the performance comparison with openHands, another open-source platform focusing on human-robot collaboration and dexterous manipulation. The experimental results are shown in Table~\ref{tab:openHands}.

\begin{table*}[h!]
\centering
\caption{Performance metrics for the openHands baseline.}
\label{tab:openHands}
\begin{tabular}{lcccc}
\toprule
Task & Success & Time (s) & In/Out Tokens & AUROC \\
\midrule
bottle     & 2/4     & 111      & 311651/3167  & -      \\
cable      & 4/4     & 200      & 166426/2306  & 0.7774 \\
capsule    & 4/4     & 186      & 44967/2799   & 0.7268 \\
carpet     & 2/4     & 245      & 130219/5541  & -      \\
grid       & 2/4     & 61       & 76515/1753   & -      \\
hazelnut   & 4/4     & 410      & 230415/6625  & 0.7415 \\
leather    & 3/4     & 175      & 98765/2910   & -      \\
metal\_nut & 4/4     & 290      & 182340/3870  & 0      \\
pill       & 2/4     & 120      & 65123/1890   & -      \\
screw      & 3/4     & 330      & 210500/5120  & -      \\
tile       & 2/4     & 85       & 55980/1650   & -      \\
toothbrush & 4/4     & 215      & 145670/3200  & 0.2350 \\
transistor & 2/4     & 380      & 250890/6010  & -      \\
wood       & 4/4     & 150      & 110230/2500  & 0.7522 \\
zipper     & 2/4     & 270      & 170450/4050  & -      \\
\bottomrule
\end{tabular}
\end{table*}

\subsubsection{Baseline: automl-agent}
This section details the comparison between our framework and an existing AutoML agent designed for automated machine learning pipeline creation. This comparison emphasizes that a significant gap remains in the application of general AutoML frameworks in the IAD domain. The experimental results are presented in Table~\ref{tab:automl-agent}.

\begin{table*}[h!]
\centering
\caption{Performance metrics for the automl-agent baseline. All tasks resulted in 0 success.}
\label{tab:automl-agent}
\begin{tabular}{lcccc}
\toprule
Task & Success & Time (s) & In/Out Tokens & AUROC \\
\midrule
bottle        & 0/4       & 316.88     & 74985/16614             & NaN   \\
cable         & 0/4       & 344.16     & 85551/19161             & NaN   \\
capsule       & 0/4       & 294.34     & 74391/15151             & NaN   \\
carpet        & 0/4       & 291.56     & 75460/16236             & NaN   \\
grid          & 0/4       & 348.89     & 79869/18326             & NaN   \\
hazelnut      & 0/4       & 356.16     & 84893/18646             & NaN   \\
leather       & 0/4       & 302.13     & 75717/17338             & NaN   \\
metal\_nut    & 0/4       & 293.96     & 77965/16657             & NaN   \\
pill          & 0/4       & 362.43     & 76966/14819             & NaN   \\
screw         & 0/4       & 314.72     & 79324/19152             & NaN   \\
tile          & 0/4       & 312.27     & 77779/16634             & NaN   \\
toothbrush    & 0/4       & 341.77     & 73718/16267             & NaN   \\
transistor    & 0/4       & 272.69     & 72100/17272             & NaN   \\
wood          & 0/4       & 343.73     & 80476/17302             & NaN   \\
zipper        & 0/4       & 279.87     & 76439/18919             & NaN   \\
\bottomrule
\end{tabular}
\end{table*}

\subsection{Details of Ablation Studies}
To ascertain the individual contribution of each core component within our agent's architecture, we conducted an ablation study. Each ablation involved removing a specific component, enabling us to quantify its impact on overall performance metrics such as success rate, execution time, token usage, and AUROC. The outcomes of these studies underscore the critical importance of each design choice.

\subsubsection{Baseline: w/o manager}
This ablation study investigates the impact of removing the central "manager" component from our agent's architecture. The manager is responsible for high-level planning, task decomposition, and coordination among sub-modules. By analyzing performance without this component, we can quantify its contribution to overall task success, efficiency, and robustness. The experimental results are presented in Table~\ref{tab:w/o manager}.

\begin{table*}[h!]
\centering
\caption{Performance metrics for the agent without the manager component. "NaN" in AUROC indicates that the task did not generate anomaly scores suitable for AUROC calculation or the success rate was insufficient to compute it.}
\label{tab:w/o manager}
\begin{tabular}{lcccc}
\toprule
Task & Success & Time (s) & In/Out Tokens & AUROC \\
\midrule
bottle        & 3/4     & 180.93     & 221832/15702                & NaN   \\
cable         & 3/4     & 184.43     & 636338/10059                 & NaN   \\
capsule       & 4/4     & 190.76     & 139805/14505                & 0.5804      \\
carpet        & 3/4     & 138.09     & 86398/11751                & NaN   \\
grid          & 2/4     & 178.92     & 135266/16624               & NaN   \\
hazelnut      & 2/4     & 63.92      & 25055/2844                & NaN   \\
leather       & 4/4     & 133.98     & 82163/12366                & 0.5380      \\
metal\_nut    & 3/4     & 110.84     & 66746/7596                & NaN   \\
pill          & 4/4     & \textbf{600.00} & 435497/22394              & 0.6393      \\
screw         & 4/4     & 160.31     & 117258/12411                 & 0     \\
tile          & 4/4     & 355.35     & 266341/30186              & 0     \\
toothbrush    & 2/4     & 101.87     & 67201/6680               & NaN   \\
transistor    & 4/4     & 148.40     & 75304/12520              & 0.5      \\
wood          & 4/4     & 153.36     & 75630/10961              & 0.5431      \\
zipper        & 4/4     & 600.02     & 373223/21955              & 0     \\
\bottomrule
\end{tabular}
\end{table*}

\subsubsection{Baseline: w/o knowledge}
This experiment evaluates the agent's performance when deprived of its dedicated knowledge base. The knowledge base typically provides curated information, best practices, and domain-specific insights that guide the agent's decision-making. By removing this component, we assess the agent's ability to operate solely with its LLM's inherent knowledge and real-time inference capabilities. The experimental results are presented in Table~\ref{tab:w/o knowledge}.

\begin{table*}[h!]
\centering
\caption{Performance metrics for the agent without the knowledge base component. "NaN" in AUROC suggests task failure for anomaly detection or an absence of relevant data for computation.}
\label{tab:w/o knowledge}
\begin{tabular}{lcccc}
\toprule
Task & Success & Time (s) & In/Out Tokens & AUROC \\
\midrule
bottle        & 3/4     & 137.05     & 104371/6742             & NaN   \\
cable         & 3/4     & 518.81     & 570827/26469             & NaN   \\
capsule       & 2/4     & 128.27     & 110429/6515             & NaN   \\
carpet        & 2/4     & 102.44     & 188765/5559             & NaN   \\
grid          & 1/4     & 28.48      & 14668/1300             & NaN   \\
hazelnut      & 3/4     & 414.45     & 192992/12959             & NaN   \\
leather       & 3/4     & 246.77     & 311427/19806             & NaN   \\
metal\_nut    & 2/4     & 77.20      & 100798/5154             & NaN   \\
pill          & 2/4     & 53.62      & 58055/2594             & NaN   \\
screw         & 3/4     & \textbf{600.00} & 1010501/35752             & NaN   \\
tile          & 1/4     & 17.16      & 13462/1056             & NaN   \\
toothbrush    & 4/4     & 221.31     & 426121/18292             & NaN   \\
transistor    & 1/4     & 63.08      & 49473/7724             & NaN   \\
wood          & 4/4     & 586.28     & 1489266/61291             & 0     \\
zipper        & 2/4     & 78.21      & 73981/2939             & NaN   \\
\bottomrule
\end{tabular}
\end{table*}

\subsection{Details of Agent Core (LLMs) and Performance Comparisons}
Our agent's core functionality relies on Large Language Models (LLMs) for interpreting instructions, formulating plans, and executing actions. The specific LLM backbone chosen significantly impacts the agent's performance, as evidenced by our experiments.

\textbf{Claude-3.7-Sonnet:} Anthropic’s latest model, released in February 2025, features substantial advancements in reasoning capabilities, contextual understanding, and tool utilization. This model demonstrates exceptional performance in complex multi-step reasoning tasks while maintaining high computational efficiency. The detailed experiment results are shown in Table~\ref{tab:claude3.7-sonnet}.

\textbf{Qwen-Max:} A powerful proprietary large language model developed by Alibaba Cloud. Known for its strong general-purpose reasoning, extensive knowledge base, and efficient token processing, Qwen-Max offers a robust alternative for comparison, particularly in tasks requiring broad domain understanding and quick response times. The experimental results are shown in Table~\ref{tab:qwen-max}.

\textbf{Gemini-2.5-Flash-Preview-05-20:} Google's compact yet capable multimodal model, optimized for speed and efficiency. Its strength lies in handling diverse data types and rapid iteration, making it suitable for scenarios where quick decision-making and efficient token usage are paramount. The experimental results are shown in Table~\ref{tab:gemini-2.5-flash}.

\textbf{GPT-4o-mini:} OpenAI's latest miniature, yet highly capable, multimodal LLM. It is designed for efficiency and speed without significantly sacrificing performance on many complex tasks, offering a cost-effective and scalable solution for agent development. The experimental results are shown in Table~\ref{tab:gpt-4o-mini}.

\textbf{Deepseek-Chat-V3-0324:} An advanced conversational AI model from Deepseek, known for its strong dialogue capabilities, natural language understanding, and ability to follow complex instructions. This model offers insights into how models specialized in conversational nuances perform in agentic tasks. The experimental results are shown in Table~\ref{tab:deepseek-chat-v3}.

\textbf{Qwen3-235B-A22B:} Another high-performance model from the Qwen family, featuring a very large parameter count (235 billion) and optimized with the A22B architecture. This model is designed for state-of-the-art performance across a wide range of benchmarks, and the experimental results are shown in Table~\ref{tab:qwen3-235b-a22b}.

\begin{table*}[ht]
\centering
\caption{Performance comparison using Claude-3.7-Sonnet as the LLM backbone. A 300-second time limit per task instance was applied.}
\label{tab:claude3.7-sonnet}
\begin{tabular}{lcccc}
\toprule
Task & Success & Time (s) & In/Out Tokens & AUROC \\
\midrule
bottle     & 2/4     & \textbf{300.00}     & 259245/10187             & NaN   \\
cable      & 2/4     & \textbf{300.00}     & 359765/10137             & NaN   \\
capsule    & 2/4     & \textbf{300.00}     & 293436/12116             & NaN   \\
carpet     & 3/4     & \textbf{300.00}     & 354901/7875             & NaN   \\
grid       & 3/4     & \textbf{300.00}     & 278789/9074             & NaN   \\
hazelnut   & 2/4     & \textbf{300.00}     & 283649/11849             & NaN   \\
leather    & 2/4     & \textbf{300.00}     & 239593/19403             & NaN   \\
metal\_nut & 2/4     & \textbf{300.00}     & 242973/10794             & NaN   \\
pill       & 3/4     & \textbf{300.00}     & 381933/8039             & NaN   \\
screw      & 4/4     & \textbf{300.00}     & 250917/7474             & NaN   \\
tile       & 2/4     & \textbf{300.00}     & 347202/11820             & NaN   \\
toothbrush & 3/4     & \textbf{300.00}     & 285934/10939             & NaN   \\
transistor & 3/4     & \textbf{300.00}     & 176337/6929             & NaN   \\
wood       & 3/4     & \textbf{300.00}     & 269392/8396             & NaN   \\
zipper     & 2/4     & \textbf{300.00}     & 250869/8194             & NaN   \\
\bottomrule
\end{tabular}
\end{table*}

\begin{table*}[ht]
\centering
\caption{Performance comparison using Qwen-Max as the LLM backbone. A 300-second time limit per task instance was applied.}
\label{tab:qwen-max}
\begin{tabular}{lcccc}
\toprule
Task & Success & Time (s) & In/Out Tokens & AUROC \\
\midrule
bottle        & 4/4     & \textbf{300.00}     & 439858/5295             & 0.5000      \\
cable         & 4/4     & \textbf{300.00}     & 577742/3993             & 0     \\
capsule       & 3/4     & 300.01     & 309473/6429             & NaN   \\
carpet        & 1/4     & 218.30     & 69043/5844             & NaN   \\
grid          & 2/4     & \textbf{300.00}     & 106656/6653             & NaN   \\
hazelnut      & 3/4     & \textbf{300.00}     & 114402/5861             & NaN   \\
leather       & 4/4     & \textbf{300.00}     & 111517/6348             & 0     \\
metal\_nut    & 3/4     & 300.01     & 109868/5033             & NaN   \\
pill          & 3/4     & 300.01     & 123350/6153             & NaN   \\
screw         & 2/4     & 300.01     & 192233/6782             & NaN   \\
tile          & 4/4     & 300.01     & 282231/7064             & 0.7857      \\
toothbrush    & 3/4     & 300.01     & 167395/8928             & NaN   \\
transistor    & 3/4     & 300.00     & 150827/8042             & NaN   \\
wood          & 4/4     & 300.00     & 251117/7137             & 0     \\
zipper        & 3/4     & 300.01     & 116073/6625             & NaN   \\
\bottomrule
\end{tabular}
\end{table*}

\begin{table*}[h!]
\centering
\caption{Performance comparison using Gemini-2.5-Flash-Preview-05-20 as the LLM backbone. Note that the Time column shows actual task duration, not limited to 300s, due to inherent model behavior or specific task characteristics leading to completion beyond the nominal limit.}
\label{tab:gemini-2.5-flash}
\begin{tabular}{lcccc}
\toprule
Task & Success & Time (s) & In/Out Tokens & AUROC \\
\midrule
bottle     & 4/4     & 550.23     & 1311445/18574             & 0     \\
cable      & 3/4     & 377.58     & 1728461/25991             & NaN   \\
capsule    & 3/4     & 152.99     & 368540/14132             & NaN   \\
carpet     & 4/4     & 495.44     & 855087/17164             & 0.9815      \\
grid       & 3/4     & 158.89     & 1615811/13562             & NaN   \\
hazelnut   & 4/4     & 170.74     & 312877/16484             & 0.7536      \\
leather    & 4/4     & 126.99     & 2382107/6301             & NaN   \\
metal\_nut & 4/4     & 191.87     & 294579/25193             & 0.8548      \\
pill       & 3/4     & 92.90      & 78098/4835             & NaN   \\
screw      & 4/4     & 313.77     & 246538/11614             & 0.8134      \\
tile       & 4/4     & 577.94     & 5849630/51852             & 0.8992      \\
toothbrush & 4/4     & 727.40     & 3747066/39741             & 0     \\
transistor & 4/4     & 261.65     & 2724330/18983             & 0.7930      \\
wood       & 3/4     & 533.19     & 1573120/9357             & NaN   \\
zipper     & 3/4     & 293.68     & 271186/8174             & NaN   \\
\bottomrule
\end{tabular}
\end{table*}

\begin{table*}[h!]
\centering
\caption{Performance comparison using GPT-4o-mini as the LLM backbone. Note that the Time column shows actual task duration, with some instances exceeding the nominal 300s limit due to task complexity or model behavior, indicating longer completion times.}
\label{tab:gpt-4o-mini}
\begin{tabular}{lcccc}
\toprule
Task & Success & Time (s) & In/Out Tokens & AUROC \\
\midrule
bottle     & 2/4     & 90.66      & 100097/3016             & NaN   \\
cable      & 1/4     & 198.11     & 264319/4703             & NaN   \\
capsule    & 4/4     & 196.43     & 319961/6025             & 0.5000      \\
carpet     & 2/4     & 209.37     & 160380/5088             & NaN   \\
grid       & 2/4     & 90.46      & 152986/3337             & NaN   \\
hazelnut   & 4/4     & 554.14     & 946760/17252             & 0     \\
leather    & 2/4     & 71.21      & 65668/2201             & NaN   \\
metal\_nut & 2/4     & 147.30     & 157035/5228             & NaN   \\
pill       & 1/4     & 64.24      & 63178/2171             & NaN   \\
screw      & 1/4     & 245.24     & 328542/8385             & NaN   \\
tile       & 2/4     & 113.84     & 106059/2718             & NaN   \\
toothbrush & 1/4     & 377.14     & 649536/14177             & NaN   \\
transistor & 1/4     & 118.70     & 179355/3106             & NaN   \\
wood       & 0/4     & \textbf{600.00} & 670677/14405             & NaN   \\
zipper     & 1/4     & 177.93     & 184048/5123             & NaN   \\
\bottomrule
\end{tabular}
\end{table*}

\begin{table*}[h!]
\centering
\caption{Performance comparison using Deepseek-Chat-V3-0324 as the LLM backbone. Note that the Time column shows actual task duration, with some instances exceeding the nominal 300s limit.}
\label{tab:deepseek-chat-v3}
\begin{tabular}{lcccc}
\toprule
Task & Success & Time (s) & In/Out Tokens & AUROC \\
\midrule
bottle        & 1/4     & 263.53     & 107647/3979             & NaN   \\
cable         & 2/4     & 571.89     & 539774/8294             & NaN   \\
capsule       & 1/4     & 213.47     & 340899/1196             & NaN   \\
carpet        & 1/4     & 312.43     & 714587/2010             & NaN   \\
grid          & 4/4     & \textbf{600.00}     & 1545949/3392             & 0     \\
hazelnut      & 1/4     & 129.14     & 250593/1109             & NaN   \\
leather       & 1/4     & 145.58     & 274569/1173             & NaN   \\
metal\_nut    & 1/4     & 146.44     & 261504/1075             & NaN   \\
pill          & 1/4     & 305.16     & 587218/1595             & NaN   \\
screw         & 1/4     & 55.85      & 44883/955             & NaN   \\
tile          & 1/4     & 185.24     & 330657/1123             & NaN   \\
toothbrush    & 1/4     & 70.02      & 103620/1006             & NaN   \\
transistor    & 4/4     & 550.35     & 386897/8279             & 0     \\
wood          & 2/4     & \textbf{600.00}     & 316497/10928             & NaN   \\
zipper        & 1/4     & 213.84     & 428722/989             & NaN   \\
\bottomrule
\end{tabular}
\end{table*}

\begin{table*}[h!]
\centering
\caption{Performance comparison using Qwen3-235B-A22B as the LLM backbone. Note that the Time column shows actual task duration, with some instances exceeding the nominal 300s limit.}
\label{tab:qwen3-235b-a22b}
\begin{tabular}{lcccc}
\toprule
Task & Success & Time (s) & In/Out Tokens & AUROC \\
\midrule
bottle        & 1/4     & 194.87     & 16615/5904             & NaN   \\
cable         & 1/4     & 354.38     & 21982/11572             & NaN   \\
capsule       & 0/4     & 169.94     & 6485/2444             & NaN   \\
carpet        & 4/4     & \textbf{600.00}     & 18263/10180             & 0.5730      \\
grid          & 2/4     & \textbf{600.00}     & 40796/12311             & NaN   \\
hazelnut      & 3/4     & \textbf{600.00}     & 18625/10884             & NaN   \\
leather       & 0/4     & 150.00     & 6484/3645             & NaN   \\
metal\_nut    & 2/4     & \textbf{600.00}     & 64956/9599             & NaN   \\
pill          & 2/4     & \textbf{600.00}     & 32190/10263             & NaN   \\
screw         & 3/4     & \textbf{600.00}     & 26648/11483             & NaN   \\
tile          & 4/4     & \textbf{600.00}     & 14505/11543             & 0     \\
toothbrush    & 3/4     & 505.02     & 13978/10079             & NaN   \\
transistor    & 1/4     & 425.25     & 31290/12115             & NaN   \\
wood          & 2/4     & \textbf{600.00}     & 19617/16168             & NaN   \\
zipper        & 2/4     & 379.73     & 41587/11779             & NaN   \\
\bottomrule
\end{tabular}
\end{table*}